\documentclass[conference]{IEEEtran}
\IEEEoverridecommandlockouts
\usepackage{cite}
\usepackage{amsmath,amssymb,amsfonts}
\usepackage{algorithmic}
\usepackage{graphicx}
\usepackage{textcomp}
\usepackage{xcolor}
\usepackage[numbers]{natbib}
\usepackage{hyperref}

\hypersetup{
  colorlinks   = true,              
  urlcolor     = black,              
  linkcolor    = black,              
  citecolor    = black               
}

\renewcommand\bibname{References}  

\def\BibTeX{{\rm B\kern-.05em{\sc i\kern-.025em b}\kern-.08em
    T\kern-.1667em\lower.7ex\hbox{E}\kern-.125emX}}

\begin{document}

\title{Improving Pose Estimation through Contextual Activity Fusion}

\author{\IEEEauthorblockN{David Poulton}
\IEEEauthorblockA{\textit{School of Computer Science and Applied Mathematics} \\
\textit{University of the Witwatersrand}\\
Johannesburg, South Africa \\
1662476@students.wits.ac.za}
\and
\IEEEauthorblockN{Richard Klein}
\IEEEauthorblockA{\textit{School of Computer Science and Applied Mathematics} \\
\textit{University of the Witwatersrand}\\
Johannesburg, South Africa \\
richard.klein@wits.ac.za}
}

\maketitle

\begin{abstract}
This research presents the idea of activity fusion into existing Pose Estimation architectures to enhance their predictive ability. This is motivated by the rise in higher level concepts found in modern machine learning architectures, and the belief that activity context is a useful piece of information for the problem of pose estimation. To analyse this concept we take an existing deep learning architecture and augment it with an additional 1x1 convolution to fuse activity information into the model. We perform evaluation and comparison on a common pose estimation dataset, and show a performance improvement over our baseline model, especially in uncommon poses and on typically difficult joints. Additionally, we perform an ablative analysis to indicate that the performance improvement does in fact draw from the activity information.
\end{abstract}

\begin{IEEEkeywords}
pose estimation, computer vision, fully convolution neural networks, context fusion
\end{IEEEkeywords}

\section{Introduction}
\begin{figure*}[]
    \centering
    \includegraphics[width=16cm,height=6cm]{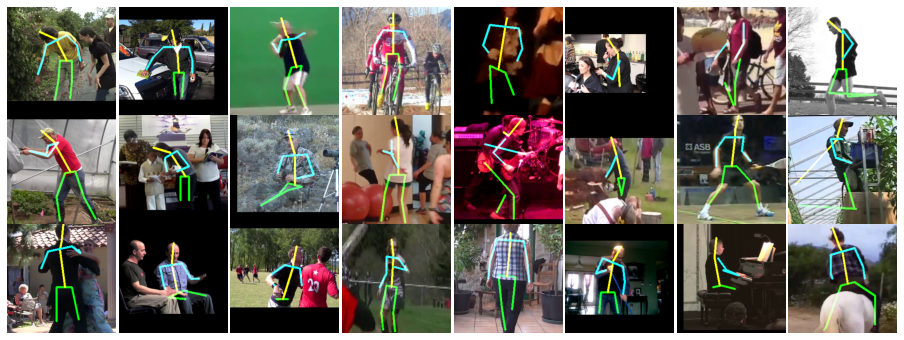}
    \caption{Some annotated examples from the test set, indicating the complexity of the problem, and the large variety of poses the model is capable of handling. The effect of having the target in the center of the image can also be noticed in images where there are several people, even in close proximity. Images are from the MPII dataset \citep{mpii_results}.}
    \label{fig:examples}
\end{figure*}

Human Pose Estimation (HPE) is a widely studied field of Machine Learning focused on finding the joint positions of humans in an image. Initially researchers developed models which functioned through the use of hand crafted features, which yielded some success, however the incredible complexity of the problem limited the viability of such methods. Some examples of pose estimation annotations can be seen in Figure \ref{fig:examples}, composed of the predictions of our final model on images randomly selected from the test set.

Over time, since the inception of neural networks, and specifically Convolutional Neural Networks (CNNs), models began to shift towards fully learned knowledge without the need for human crafted features or prior information. CNNs enabled models to effectively handle image data due to the nature of their design, which suited HPE and propelled model accuracies and speeds, with networks becoming larger and more carefully structured. Architectures such as DeepPose \citep{toshev2014deeppose} and Stacked Hourglass \citep{newell2016stacked} were forerunners of deep learning for HPE.

The more complex structures that fell under deep learning began enabling models to capture higher level concepts such as whole objects and bodies, above local features or individual limbs \citep{toshev2014deeppose,rafi2016efficient,tompson2015efficient}, improving their accuracies. This indicates that modern architectures are capable of utilising these concepts, and designing them with this in mind can yield high performing models. Following on from that, we predict that the use of contextual information within pose estimators can further improve performance.

In humans, different activities generally tend to contain markedly different poses, and in turn different poses tend to be found more amongst certain activities, and so we choose activity as our contextual information of choice. Our goal here is to determine whether knowing an image contains a certain activity can drive the model to certain biases that improve its ability to estimate poses. For example, knowing an image comes from a rugby game may inform a pose estimator that it will likely encounter more running, diving, and kicking, rather than sitting.

However, we also need to consider how images are categorised to activities, which is not straightforward. Using one set of activities may be more reasonable conceptually, where using another may better segment poses but be less useful. For example, using ``Playing soccer'' is more practical than decomposing it into ``Running'', ``Kicking'', and ``Tackling'', even though the latter decomposition may better separate the poses we expect to see.

Even finding an effective way to fuse this activity information into the current deep learning networks available is not a trivial design choice due to the complex nature of modern architectures. Above that, where to provide this information to models requires some thought as well. We hope to address at least some of these concerns through this research.

The remainder of the paper is structured as follows: Section \ref{sect:related_work} explores some of the existing work in the field of pose estimation, as well as related concepts of context fusion in imagery. Section \ref{sect:architecture} then covers the architectural design choices for the research. Section \ref{sect:training} provides detail on how our chosen models were trained, followed by the results thereof in Section \ref{sect:results}, including an ablative analysis of our method. We then provide some concepts for future work in Section \ref{sect:future_work}, followed by our conclusion in Section \ref{sect:conclusion}.

\section{Related Work}\label{sect:related_work}

Early models of HPE utilised largely parts based models \citep{shakhnarovich2003fast,yang2011parts}, which were successful at the time but fall well short of the accuracies enjoyed by modern deep learning networks, and are not well suited to higher level conceptual learning.

\begin{figure*}[]
    \centering
    \includegraphics[width=18cm,height=6cm]{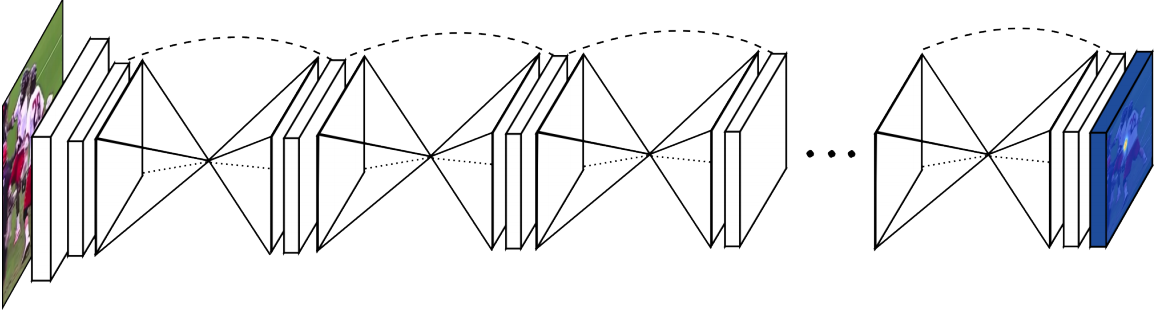}
    \caption{A visualisation of the original Stacked Hourglass architecture, taken from \citet{newell2016stacked}.}
    \label{fig:hourglass_original}
\end{figure*}

Modern networks are far better suited to the task, and fall largely into one of two categories, namely regression based or heatmap based models. DeepPose \citep{toshev2014deeppose} was a forerunning regression based model. It utilised a holistic method to determine initial pose estimates, and then followed up with a series of cascading CNNs to refine predictions on a per-joint basis.

On the other hand, models such as the Stacked Hourglass model \citep{newell2016stacked} utilise heatmaps entirely for joint predictions. The Hourglass model takes advantage of contractions and expansions, as well as residuals, to find a good balance between holistic, global features, and smaller, localised details in images. Wei et al. \citep{wei2016convolutional} developed a similar concept of a repeated sequence of sub-CNN modules which each produce a heatmap that is passed on to the next module, inspired by \citep{toshev2014deeppose}. This structure also displayed improvements in the ability of the model to capture both global and local contextual pose information.

Some approaches also utilised model-based learning to restrict pose estimates to realistic spaces. \citet{sun2017compositional} implemented a bone-based representation that allowed for learning of skeletal structure, rather than directly predicting joint positions. They also adapted their loss functions to account for errors specific to joints, and errors caused by misalignment of bones between the root and current joint. Both these alterations successfully increased accuracy over model-free methods seen before. \citet{bourached2020generative} utilise a model-based generative architecture to try improve out-of-distribution poses, enhancing the models ability to generalise, rather than focus on more accurate predictions for known distributions.

Rather than using repeated stages of differing refinements, \citet{chen2017adversarial} attempt to ensure realistic poses by utilising a GAN system, where adversaries are used to determine how reasonable the generated pose and confidence maps are. An ablative study indicates the GAN structure is indeed contributory to accuracy improvements. \citet{bin2020adversarial} also utilise a GAN network to augment inputs and make them easier to predict on, yielding the best performance achieved on the MPII dataset.

As for the concept of activity fusion, it appears that the available literature is relatively sparse. \citet{rafi2015semantic} explores the concept of semantic and contextual information by utilising a depth-based pose estimator which is capable of identifying objects in the scene and using them for context, however this only used a small set of objects. \citet{Vu_2015_ICCV} also uses contextual scene information in its estimation process which improved performance, however the model is only applicable for head estimations, whereas our focus is on full body poses.

Utilising depth information in their approach, \citet{JALAL2017295} extracts skeletal information in order to produce an activity estimate, however their activity classifications are much finer grained than those utilised here, using classes such as walking or sitting rather than exercising. Rather than depth, \citet{Gowda_2017_CVPR_Workshops} uses RGB imagery and extracts poses to be used for broader activity estimates, which is more relevant to the topic, and indicates there is a possible relationship of significance between pose and activity.

\citet{LAN2020156} explores a similar concept of utilising contextual information around an image, however in their use case it is applied to road image segmentation rather than human poses. Their approach still yielded favourable results, and is encouraging for our concept.

\section{Architecture}\label{sect:architecture}

The utilised architecture was based off of the Stacked Hourglass model \citep{newell2016stacked} with some added layers for the activity fusion. The Stacked Hourglass model was selected because of its high base accuracy, inherently flexible modular design, and balance at finding global and local pose cues. It also maintains the same shape of features throughout the majority of the model, specifically $64 \times 64 \times 256$, making testing various fusion sites more straightforward.

Because the Hourglass network is already designed with capturing global context in mind, it was also of interest to see if explicitly providing the context would have a significant impact, or if the model itself was already capable of extracting the context in some sense.

We use the final version of the Stacked Hourglass model initially presented in \citep{newell2016stacked} as our baseline, composed of eight glasses, and utilising intermediate supervision for training. A rough visualisation of the original network can be seen in Figure \ref{fig:hourglass_original}.

Our baseline model still makes use of the initial down-convolution segment of the network, as well as the intermediate bottlenecks and final remapping convolutions, and regularisation and frequent batch normalisation. We then had three sets of models, each composed of an ablative model and a contextual model. The ablative models were the same as our baseline, however with a single extra one-by-one convolution inserted at a specific point in the network. The contextual model involved stacking the activity tensor on top of the existing tensor in the model at the point, followed by a one-by-one convolution, yielding the so called fused image that is then propagated normally through the remainder of the network.

\begin{figure}[]
    \centering
    \includegraphics[width=6.5cm,height=3cm]{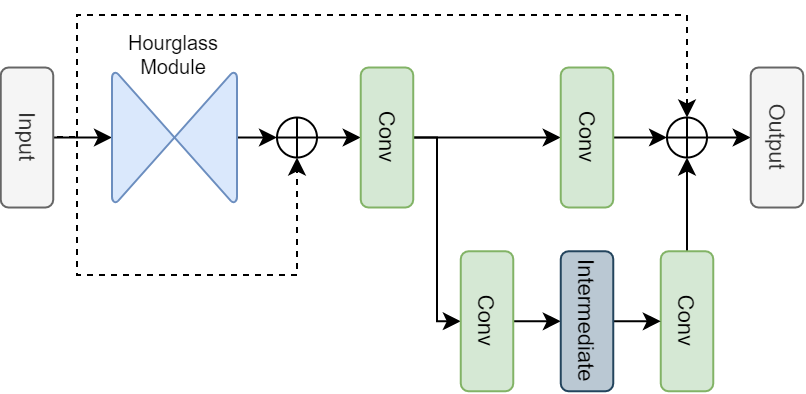}
    \caption{The original hourglass block format. The input to the module comes from previous layer of the network, and the block produces output for the next block as well as an intermediate heatmap prediction.}
    \label{fig:original_module}
\end{figure}

\begin{figure}[]
    \centering
    \includegraphics[width=8cm,height=3cm]{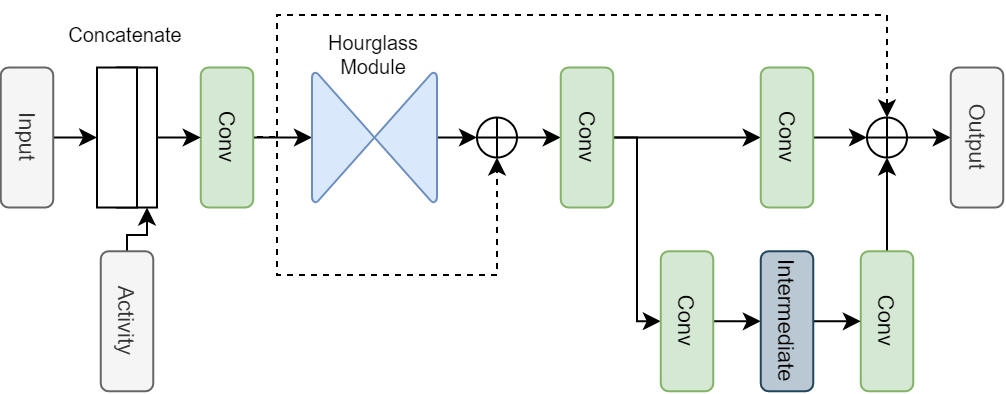}
    \caption{The context hourglass block used for our fusions. The input to the module comes from the stacking of the activity tensor and the previous layer in the regular network. The ablative version leaves out the activity tensor stacking.}
    \label{fig:context_module}
\end{figure}

We make use of the one-by-one convolution in order to correct the number of layers in the tensor after we have stacked our activity onto the previous output. Throughout the majority of the network there are 256 channels in the output of the layers, and so stacking our 21 channel activity tensor brings the tensor to 277 channels. In order to minimise the changes needed to the network, we then convolve the tensor back down to 256 layers. This method also ideally allows the convolution to learn an effective mapping to merge our context into the tensor, without having to rely on the existing layers in the network. A diagram of this context block can be seen in Figure \ref{fig:context_module}, compared to the original hourglass block in Figure \ref{fig:original_module}.

We utilise both an ablative and contextual version to verify that any possible changes in the accuracy of the model are due to the impact of our activity fusion, rather than the increase in the size of the model over the baseline. Any increase in size or alteration to the flow of the network may be significant enough to noticeably improve the network's ability to learn features, regardless of whether we provide activity context or not. Only testing our contextual augmentations would not reveal the source of improvement. This means we need to test both the contextual augmentation, as well as an ablative version without the activity, to determine if any accuracy changes are resultant from the one-by-one convolution itself, from the contextual information, or from both. If our contextual augmentation outperforms our ablative model, we can have confidence that the improvement is owing to the context itself.

\begin{figure*}[]
    \centering
    \includegraphics[width=16cm,height=4.5cm]{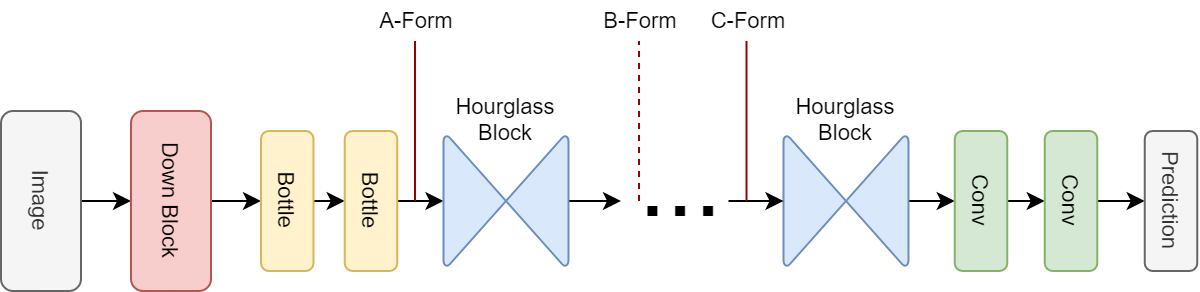}
    \caption{The augmented network, with lines indicating the three tested fusion points. The input to the fusion modules come from the activity tensor and the previous layer before the insertion point.}
    \label{fig:hourglass_augmented}
\end{figure*}

For simplicity we refer to the three different augmented networks as A-, B-, and C-Form, lettered from left-to-right according to their injection points, as can be seen in Figure \ref{fig:hourglass_augmented}. A-Form has the fusion before the first hourglass, B-Form before the fourth, and C-Form before the eighth. Their matching ablative models are referred to as A-, B-, and C-Form Ablative. Each line in the figure indicates a position where we tested our augmentation. Note that only a single fusion point was used at a time for each model.

The decision of how to fuse the activity was made difficult by the convolutional nature of the network, which stopped us from simply concatenating on a one-hot encoded vector to our input. To get around this our context took the form of a one-hot encoded activity tensor of size $64 \times 64 \times 21$, where 21 is the number of activities present in the dataset, and $64 \times 64$ is the shape of the image features as they move through the network.

\section{Training}\label{sect:training}

Training was performed in a similar manner to that of the original paper \citep{newell2016stacked} for the MPII Human Pose dataset \citep{andriluka14cvpr} only. The MPII dataset provides approximately 25k images (2.5k of which are withheld as their own hidden test set), leaving 22.5k images for use. Of those, approximately 18k (80\%) were used for training, with the remaining split evenly between a validation and test set of 2.25k (10\%) each.

The dataset provides a 2D ground-truth pose with joint visibility flags, an approximate scale, and an approximate center position, among other data not relevant to our analysis, for each person within an image. Images also have an activity classification which is used for our context, which many other datasets do not provide, making MPII particularly valuable. While MPII provides more fine grained activity sub-classifications than the 21 activity categories utilised, we did not use them as our context tensors would become intractably large, and many of the sub-classifications had few or no samples.

Because images can contain more than one person, we follow the method of \citep{newell2016stacked} by cropping the image around the center of the target so the model knows which person to estimate on, The images are cropped based on the provided scale and the provided center coordinates of the target, and the crop is then scaled to $256 \times 256$. Cropped images also undergo some additional rotation (in $[-30, 30]$) and scaling (in $[0.75,1.25]$) to provide augmentation to the dataset.

One issue encountered with the dataset was that utilising the scale provided for each person did not always result in the full pose being in frame. The scale value indicates a box of width 200px around the center of the target, however even using $1.5\times$ the scale occasionally resulted in extremities such as ankles or wrists not appearing in the final crop, which may lead to a decrease in accuracy and increases the difficulty of identifying the already challenging joints. Regardless, utilising the scale value as-is has been the approach taken by several papers that utilise the MPII dataset \citep{newell2016stacked, bin2020adversarial, yang2017learning, tang2018quantized}, and so we follow the same approach.

The cropping and augmentation can also result in black pixels appearing in the final image provided to the network, as black background pixels can be included from outside the bounds of the image. Obviously this is not very realistic, however the complex networks are likely capable of learning to ignore this anomaly. As a possible alternative, mirror padding could be used when cropping, however, again, the simple background inclusion is the approach taken by several papers \citep{newell2016stacked, bin2020adversarial, yang2017learning, tang2018quantized}, and so we use the same method.

The baseline model was provided with images cropped around the center of the person with small random rotations and scaling modifications, and a heatmap was generated for each joint of the target pose as ground-truth for the model. The augmented models were provided both with the image crop, and the one-hot encoded tensor for the activity representation.

All models were trained with an MSE loss function over the ground-truth heatmap and predicted heatmap set. The model makes heavy use of batch normalisation \citep{ioffe2015batch} to improve training speed and performance. This allowed us to use a relatively high learning rate of $2.5^{-4}$ with the RMSprop optimiser, taken from the original Stacked Hourglass paper \citep{newell2016stacked}. We utilised our validation set to select the best model as that with the lowest validation loss at any point of training.

Training of the models took approximately three days using an RTX3080 and a batch size of 8, which was the largest we could achieve given the complexity of the network. This meant a prediction time of ~60ms per image. This was for both our baseline model and our augmented model set, indicating that our context pipeline has a negligible overhead.

\section{Results}\label{sect:results}

While it does not imply an increase or decrease in accuracy, it should be noted that results were evaluated using our own withheld test set comprised of the released annotations, not the official MPII test set. This is due to the now publicly available test set not having associated activities, and attempts at getting activity annotations being unsuccessful. We utilised the common PCKh@0.5 metric, which represents the percentage of correct keypoints, where a correct keypoint is defined as being within a threshold, specifically half the normalised distance between the top of the neck and the top of the head, of the ground-truth keypoint.

\subsection{Evaluation}

Our final total accuracy for the best model, the C-Form model, was 90.3\%, an improvement of 2.7 percentage points over our baseline model's performance of 87.6\%. The contextual information was particularly impactful on some conventionally difficult joints such as ankles, knees, and wrists, where the contextual model improved over the baseline by 7.6\%, 3.3\%, and 1.5\% respectively. The accuracies of the different models can be seen in Table \ref{tab:contextual_accuracies}.

\begin{table*}[]
    \centering
    \caption{Joint and total accuracies of different networks when run on our own test set.}
    \renewcommand{\arraystretch}{1.3}
    \begin{tabular}{|l|lllllllllll|}
    \hline
    \textbf{Model} &
      \multicolumn{1}{l|}{\textbf{Head}} &
      \multicolumn{1}{l|}{\textbf{Neck}} &
      \multicolumn{1}{l|}{\textbf{Torso}} &
      \multicolumn{1}{l|}{\textbf{Pelvis}} &
      \multicolumn{1}{l|}{\textbf{Shoulder}} &
      \multicolumn{1}{l|}{\textbf{Elbow}} &
      \multicolumn{1}{l|}{\textbf{Wrist}} &
      \multicolumn{1}{l|}{\textbf{Hip}} &
      \multicolumn{1}{l|}{\textbf{Knee}} &
      \multicolumn{1}{l|}{\textbf{Ankle}} &
      \textbf{PCK@0.5} \\ \hline
    \textbf{Baseline} & 91.6 & 97.2 & 98.0 & 92.7 & 93.5 & 91.1 & 88.7 & 88.7 & 83.2 & 55.9 & 87.6 \\ \hline
    \textbf{A-Form} & 93.5 & \textbf{98.1} & 99.1 & 93.7 & 95.3 & 92.1 & 90.1 & 90.8 & 86.0 & 58.8 & 89.4 \\ \cline{1-1}
    \textbf{A-Form Ablative} & 92.8 & 97.7 & 98.8 & 94.2 & 94.6 & 91.7 & 89.9 & 90.6 & 86.4 & 60.9 & 89.3 \\ \hline
    \textbf{B-Form} & \textbf{94.0} & 97.9 & \textbf{99.3} & 94.5 & 95.1 & 92.4 & 90.1 & 91.3 & \textbf{86.9} & 63.3 & 90.0 \\ \cline{1-1}
    \textbf{B-Form Ablative} & 92.2 & \textbf{98.1} & 99.0 & 94.2 & 95.0 & 92.5 & 89.9 & 90.7 & \textbf{86.9} & 60.8 & 89.5 \\ \hline
    \textbf{C-Form} & 93.7 & 98.0 & 99.2 & \textbf{95.3} & \textbf{96.0} & \textbf{92.7} & \textbf{90.2} & \textbf{91.7} & 86.5 & \textbf{63.5} & \textbf{90.3} \\ \cline{1-1}
    \textbf{C-Form Ablative} & 92.4 & 96.1 & 97.1 & 94.1 & 93.2 & 88.0 & 85.2 & 88.4 & 78.8 & 56.8 & 86.2 \\ \hline
    \end{tabular}
    \label{tab:contextual_accuracies}
\end{table*}

\begin{figure}[]
    \centering
    \includegraphics[width=8cm,height=5cm]{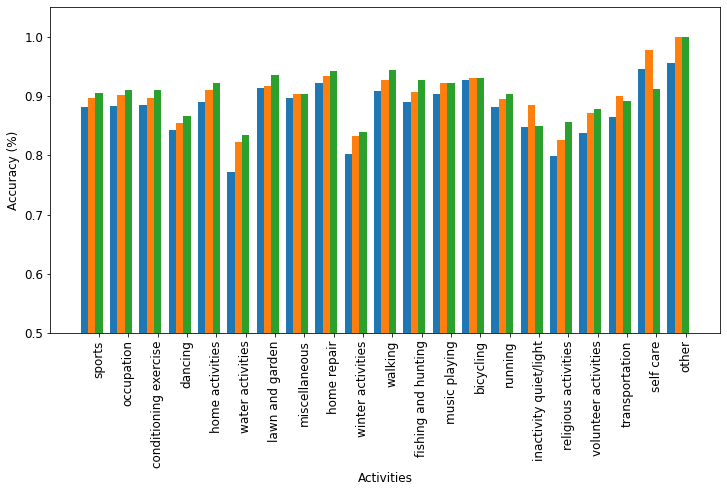}
    \caption{The different activity accuracies for our baseline (blue), B-Form Ablative (orange), and best augmented model, the C-Form model (green). Activities are sorted left to right by the number of images in the test set, ranging from 600+ (leftmost) to fewer than 10 (rightmost).}
    \label{fig:activity_accuracies}
\end{figure}

\begin{figure*}[]
    \centering
    \includegraphics[width=12cm,height=6cm]{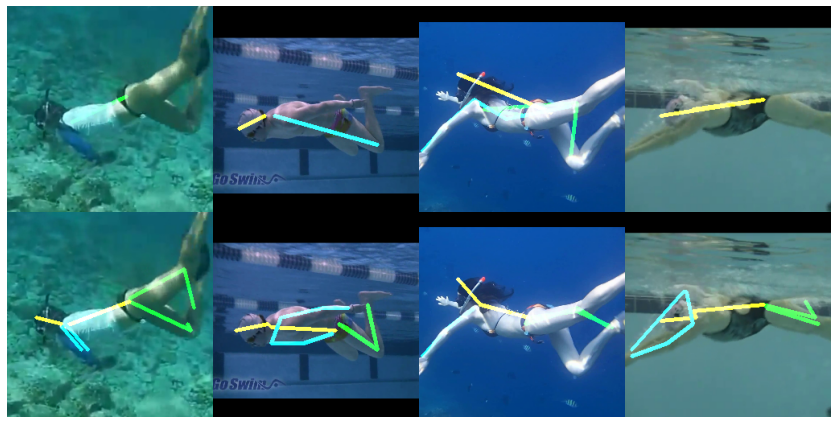}
    \caption{Images from our test set showcasing how the C-Form model (bottom) is better capable of handling the relatively uncommon poses that fall under the ``Water Activities'' classification, compared to the Baseline model (top).}
    \label{fig:water_examples}
\end{figure*}

In terms of per-activity accuracies, our C-Form model showed noticeable improvements on most activities, faring better in activities with more data points. The average activity improvement was 2.5\%. The apparent variance in improvements per activity is likely due to the nature of the activities and their compilations. For example, our model saw the largest improvement in the ``Water Activities'' classification of 6.3\%, where the poses are very different from those found in other activities, with some subjects being in unusual stances, obscured in scuba gear, or even upside down. A comparison on this activity between the Baseline and C-Form model can be seen in Figure \ref{fig:water_examples}. The ``Miscellaneous'' and ``Bicycling'' classifications on the other hand showed minimal improvement, likely due to the relatively random grouping, and already common poses that require minimal context, respectively. The different per-activity accuracies can be seen in Figure \ref{fig:activity_accuracies}.

\begin{figure}[]
    \centering
    \includegraphics[width=6cm,height=5cm]{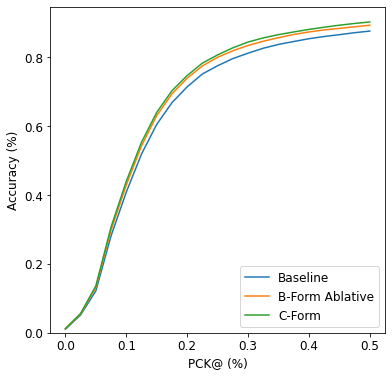}
    \caption{The different accuracies when varying PCKh thresholds for our baseline, best ablative model, and best contextual model}
    \label{fig:pckh_accuracies}
\end{figure}

Our C-Form model also shows consistent improvement over the baseline when evaluating at varying PCKh thresholds, having a higher accuracy at every value. This indicates that the performance improvement is actually caused by overall better predictions, rather than by chance that a set of keypoints moved within the threshold distance only for a specific value. This can be seen in Figure \ref{fig:pckh_accuracies}.

This indicates that our method of activity fusion may very well be useful if our activities are structured well enough to segment very different poses and cluster similar ones. Naturally this is in itself a challenging task, but nevertheless means there may be room for improvement.

Interestingly, our results indicate some level of sensitivity of the method to the fusion position. Our ablative models experienced a sharp decline in performance towards the end of the model, with the C-Form actually performing worse than our baseline. Alternatively, our contextual models performed progressively better with fusions towards the end of the model. Only providing context at the beginning of the network may be too early to enable the model to effectively utilise the information for the final predictions, with the signal from the context having diminished by the time it reaches the later hourglasses. 

In comparison to state of the art methods, our model seems to perform slightly below the state of the art accuracies reported on the MPII test set \citep{mpii_results}. Again, this is not an entirely useful comparison however, as our research was unable to make use of the official test set, and so our true test result is unknown. Nevertheless, the intent of the research was to indicate the usefulness of activity fusion, which was apparent in our noticeable accuracy improvements.

\subsection{Ablative Experiments}

All of our contextual models fairly significantly outperformed the baseline model in their overall accuracy, however we needed to perform an ablative test to ensure the contextual information itself is actually contributory to our results.

Our testing indicated that adding the additional one-by-one convolutions does, in some cases, increase the performance of the network, namely the A-Form Ablative and B-Form Ablative models. However, performance degrades in the C-Form Ablative model, and so clearly the increase in the C-Form contextual model's accuracy is not only attributable to the additional convolution.

The best ablative model, the B-Form Ablative, showed a performance improvement of 1.9\% over the baseline. While such a margin may simply be caused by variability in training, the consistent trend amongst the different augmentation sets seems to indicate at least some level of impact by both the architectural changes, as well as the provision of context. The overall accuracies for each model can be seen in Table \ref{tab:ablative_accuracies}.

\begin{table}[]
    \centering
    \caption{Total accuracies for each model on our own test set.}
    \renewcommand{\arraystretch}{1.2}
    \begin{tabular}{|l|l|}
        \hline
        \textbf{Model}  & \textbf{Accuracy} \\ \hline
        Baseline        & 87.6              \\ \hline
        A-Form          & 89.4              \\
        A-Form Ablative & 89.3              \\ \hline
        B-Form          & 90.0              \\
        B-Form Ablative & 89.5              \\ \hline
        C-Form          & \textbf{90.3}              \\
        C-Form Ablative & 86.2              \\ \hline
    \end{tabular}
    \label{tab:ablative_accuracies}
\end{table}

It should be noted that the C-Form Ablative model also consistently performed worse than the baseline model, which is unusual as the model could simply learn an identity mapping and have no affect on our predictions. This drop in accuracy could possibly be due to the degrading of gradients during back-propagation, as the Stacked Hourglass makes extensive use of residuals, and adding extra convolutions without skip layers may hinder this.

\section{Future Work}\label{sect:future_work}

While we have explored the basic concept of activity fusion in this report, there are already numerous apparent avenues to continue analysing in various domains.

In terms of the model utilised, this paper focused on a relatively conceptually simple deep learning model, however in the future we could rather make use of even newer state-of-the-art performers. Additionally we could design our own network centered around the concept of activity fusion.

We can also perform more rigorous testing of our results, performing cross-fold validation to account for the official test set not being usable, which would eliminate the concern of random variance. Furthermore we could analyse how our activity affects the features our model learns and makes use of, should it have a significant impact.

Exploring more methods of encoding and fusing our context would also be useful, as we only focused on straightforward one-hot encoding with concatenation and a $1 \times 1$ convolution to fuse the information. Alternatives could involve using auto-encoders before providing the input to the network, or other latent space methodologies. We could also investigate different methods of fusing involving a more thorough merging strategy rather than just a single convolution layer.

Finally we could explore the usefulness of context in other deep learning fields. While pose estimation seems like a useful field for context provision, there may be many others, such as image segmentation, or translation systems. 

\section{Conclusion}\label{sect:conclusion}

In this paper we explored the concept of fusing contextual activity information into the existing Stacked Hourglass model. We show that even with rudimentary organising of images into activities, and using straightforward fusing methods, our method is capable of providing performance gains over a baseline model. Above this our method introduced no significant overhead into the training or prediction process.

Our method was capable of improving accuracy on typically difficult joints, and is especially useful in activity classifications where poses are unusual in comparison to the available training data. We also provide various avenues for further exploration, and are hopeful that context fusion is a viable addition to improving deep learning models in the field and beyond.

\section*{Acknowledgments}

We acknowledge the Centre for High Performance Computing (CHPC),
South Africa, for providing computational resources to this research project. Additionally, this work is based on research supported in part by the National
Research Foundation of South Africa (Grant Numbers: 118075 and 117808).

\bibliography{references}

\end{document}